\documentclass{article}

\usepackage{graphicx}
\usepackage{amssymb,amsmath}

\usepackage[linesnumbered,ruled,vlined]{algorithm2e}

\title{Generate and Verify \\Semantically Meaningful Formal Analysis of Neural Network Perception Systems}
\author{
  C. R. Serrano, P. M. Sylla, M. A. Warren}
\date{\small Information and Systems Sciences Laboratory\\
  HRL Laboratories LLC\\Malibu, CA}

\begin{document}
\maketitle

\begin{abstract}
  Testing remains the primary method to evaluate the accuracy of
  neural network perception systems. Prior work on the formal
  verification of neural network perception models has been limited to
  notions of local adversarial robustness for classification with
  respect to individual image inputs. In this work, we propose a
  notion of global correctness for neural network perception models
  performing regression with respect to a generative neural network
  with a semantically meaningful latent space. That is, against an
  infinite set of images produced by a generative model over an
  interval of its latent space, leverage neural network verification
  tools to prove that a perception model will always produce estimates
  within some error bound of the ground truth. Where the perception
  model fails these tools will return a semantically meaningful
  counter-example. These counter-examples are semantically meaningful
  in the sense that they carry information on concrete states of the
  system of interest that can be used programmatically without human
  inspection of corresponding generated images. Our approach, Generate
  and Verify, provides a new technique to gather insight into the
  failure cases of neural network perception systems and provide
  meaningful guarantees of correct behavior in safety critical applications. 
\end{abstract}

\section{Introduction}
One would often like to know whether or not a neural network
perception system produces correct estimates and, moreover, when it
will fail to do so. The na\"ive approach to this question is to
gather more data against which to evaluate system performance. This
gives a necessarily incomplete picture:
we can only ever evaluate a finite set of individual data points
(e.g., images), and, in
most cases, it will be impossible to meaningfully capture the totality
of relevant data the system might encounter in its operational
environment by such a set. Indeed, part of the appeal of neural
networks and machine learning is their ability to, in many cases,
successfully generalize to unseen data. 

\begin{figure}
	\centering
	\includegraphics{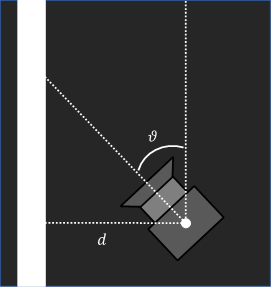}
	\caption{Example of configuration space for simple distance to
		line estimation. Here $d$ is the distance in meters to the
		line and $\theta$ is the yaw angle.}
	\label{fig:cspace}
\end{figure}

A recent alternative approach utilizes neural network verification
tools built on techniques such as satisfiability modulo theories
(SMT), mixed integer linear programming (MILP), or other techniques
from logic and computational geometry to mathematically analyze the
neural network and produce either proof artifacts, capturing
mathematical guarantees on performance, or
counter-examples, indicating cases where mathematical guarantees are
impossible. Crucially, because of the analytical approach of neural
network verification, the guarantees obtained are universally
quantified and therefore apply to an entire region of the input space
of the network encompassing infinitely many input values.  The
skeptical reader should compare this with cases such as linear
programming, where (in infeasible instances) similar guarantees over
infinite sets are possible.  Neural network verification promises to powerfully
complement testing, but it has primarily been leveraged in
the perception domain in connection with \emph{local adversarial robustness}
properties of the form:
\emph{Given a model and fixed image $x$, does there exist a
	perturbation $\delta$ bounded in norm below some fixed
	constant $\epsilon > 0$ such that the perturbed
	image $(x + \delta)$ is classified differently from $x$ by the model?}

Beyond local adversarial robustness and related properties, such as
invariance under specified transformations, the use of neural network
verification to analyze perception systems is a challenge, due to the
fact that most of the desirable properties of
such systems cannot be formally specified. Having a formal,
mathematically well-defined, specification is a prerequisite for the
use of these techniques.  E.g., how would one specify the property
that a classifier will correctly classify pedestrians in a
formal, and non-statistical/empirical, mathematical manner?  It is
apparent that the formal specification of such properties in a way that
can be analyzed by neural network verification is impossible.
Nonetheless, our goal in this paper is to show that, by the use of
generative models (which introduce an empirical/statistical component
to the analysis), it is possible to use neural network verification
techniques to specify and formally reason over more general
model properties than local adversarial robustness and transformation
invariance.

If we are interested in global correctness with respect to
semantically meaningful specifications, then the images over which we
reason will need to be drawn from a semantically meaningful
latent space. Our approach to global correctness first defines a
semantically meaningful \emph{configuration space} capturing the
properties of the scenario depicted in images of the kinds we wish to
reason over. As a motivating example, our work has focused on the
analysis of a perception system for estimating the distance to a lane
marker from images produced by a front facing vehicle mounted
camera. The configuration space consists of the vehicle's distance to
the line $d$, and yaw angle $\theta$ with respect to the line, as
depicted in Figure \ref{fig:cspace}. 

To then verify properties of a neural network perception system with
respect to our configuration space we require a function (the
\emph{decoder}) mapping from
configuration space to image space, which can then be
composed \emph{in-line} with our regression perception system
(the \emph{regressor}). The formal analysis using neural network
verification then reasons over a semantically meaningful region of the
configuration space, and the output to be checked for correctness is
that of our regressor. Each interval in our configuration space,
assuming it is not degenerate, represents an infinite set of images
against which the perception system will be verified for correctness. 

Our approach, \emph{Generate and Verify}, learns a neural network decoder
from configuration space to image space against which the regressor
can be verified. Our notion of global correctness is thus modulo the
learned decoder. In the event the neural network verification cannot
prove correctness it will generally return a counter-example in configuration
space, which can then be provided to the decoder to yield the exact
image on which our perception system fails. In practice, we divide the set of intervals
over our configuration space into smaller boxes so that we may
parallelize computation, gather diverse counterexamples, and identify
smaller regions over which our regressor may be provably correct. Our
approach yields proof artifacts (certificates), and provides a method to find
counter-examples, that traditional sample and test techniques cannot provide.  

\begin{figure*}[t]
	\centering
	\includegraphics[width=0.75\textwidth]{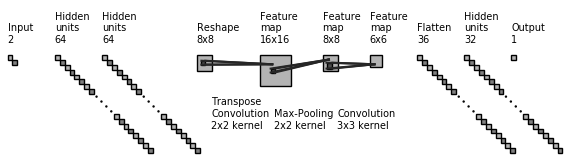} % Reduce the figure size so that it is slightly narrower than the column.
	\caption{The combined decoder and regressor architecture. The decoder maps $C$ space input to the 16x16 intermediary image space $I$ while the regressor maps from $I$ to a single output $\in \mathbb{R}$.}
	\label{fig:architecture}
\end{figure*}

\section{Related Works}
Our work is closely related to prior work on the verification of
neural networks \cite{Katz2017ReluplexAE, Katz2019TheMF,
  Tran2020VerificationOD, Tran2019StarBasedRA, Dutta2019SherlockA,
  Singh2019AnAD, Singh2019BeyondTS, Tjeng2019EvaluatingRO,
  Balunovic2019CertifyingGR,Botoeva2020EfficientVR}. The
setting for the tools developed in those works is as follows. Assume
given an input set $X$ (e.g. image space) and an output set $Y$ where
elements of $Y$ correspond to either different discrete classes or
probability vectors over classes. The neural network itself is
mathematically regarded as a function $f\colon X \rightarrow Y$ and the
local robustness properties considered in the works above state that,
for a fixed $\epsilon > 0$ and input $x$, there exists no perturbation
$\delta$ with $||\delta|| < \epsilon$ (in a fixed norm, usually
$L_\infty$) such that $f(x)$ and $f(x + \delta)$ are too far from one
another. Minor variations on this setting, where $f(x + \delta)$
should not be too far from the true class, are also considered. This
is a reasonable property to consider that directly relates to the
phenomena of adversarial examples, but it tells us nothing about
the kinds of images more generally on which the neural network will be
likely to correctly perform. Instead, it merely bounds how far small
perturbations $\delta$ of \emph{fixed images} are able to push the
resulting output of the network.

A notable exception to the literature above is the technical report
\cite{Fijalkow2019VerificationON} in which an approach to global
correctness obtained by pre-composing a generative model with an image
classifier to find counter-examples through random search is
described. Crucially, this work relies on the
sampling of noise from a (semantically meaningless) latent space (of
noise) to generate \emph{finite} sets of images against which a
classifier is tested.  By contrast, our work first constructs a
semantically meaningful space from which images are generated, and
evaluates model correctness, using neural network verification,
against \emph{infinite} sets of images. 

A final paper that is relevant to our work, but rather distinct, is
\cite{Fremont2019ScenicAL} which introduces a formal language for
specifying scenes and a compiler for rendering scenes in a
pre-existing simulator. This approach has been used in conjunction
with targeted testing of neural network vision systems. However, as
with traditional testing, the outputs of this approach are the same as
traditional sample and test approaches: individual instances of
failure or success independent of any global correctness guarantee.

\section{Background}

\subsection{Neural Network Verification}
Our approach is largely agnostic regarding the choice of neural
network verification tool: as long as the tool is capable of reasoning
over the decoder and regressor it can be used.  In the experiments
described below, we used the general purpose \emph{dReal} SMT solver
\cite{gao2013dreal}.  The advantage of dReal is that, by virtue of the
so-called ``$\delta$-decidability theorem'' of \cite{Gao:delta}, dReal is able to
handle reasoning over properties involving non-linear real arithmetic
and all of the common non-linear functions (including transcendental
functions).  At a very high level, dReal transforms problems of
real arithmetic with non-linear functions into the domain of interval
analysis and then carries out a DPLL-style \cite{Selman:SAT}
branch-and-prune search by splitting boxes until a fixed precision
(the ``$\delta$'') has been reached.  Consequently, we are able to reason
over neural networks with all common neural network architecture
features: convolutions, transposed convolutions, ReLU, tanh, sigmoid,
softmax,  batch normalization, etc., can all be represented and
reasoned over in dReal.  In addition to dReal itself, we make use of a
custom C++ library for transforming neural network models into a
format that dReal can handle (dReal was designed to reason over
scalars and our library can be regarded as a wrapper on dReal at the
level of tensors).

In terms of usage, dReal is provided with a list of variables
$x_{1},\ldots,x_{n}$ over which we intend to solve.  We provide also a
property $\varphi(x_{1},\ldots,x_{n})$ to be evaluated.  The results
of the evaluation are of three kinds.  First, when the property is
satisfiable (\emph{SAT}) dReal will provide us with a list
$a_{1},\ldots,a_{n}$ of concrete values such that
$\varphi(a_{1},\ldots,a_{n})$ holds.  When the property is
unsatisfiable (\emph{UNSAT}) dReal provides us with a mathematical
proof (certificate) establishing that the property holds for non values of
$x_{1},\ldots,x_{n}$.  Finally, the property could be \emph{unknown}
in which dReal can be re-run with a smaller precision
parameter $\delta$, although, due to undecidability of first-order
real arithmetic with transcendental functions, we cannot guarantee
that the process of progressively decreasing the magnitude of
$\delta$ and re-running dReal will result in a SAT or UNSAT.

The principal bottleneck to the broad use of neural network
verification tools is scalability, although recent advances such as
\cite{Tran2020VerificationOD} are promising.  With this in mind, we
describe timing data on the use of the verification tools in detail
below.  It is worth noting that one advantage of our approach is that
it moves the problem of neural network verification from
high-dimensional image space, where these tools scale more poorly, to
the lower dimensional spaces $C$ and $C\times Z$, where these tools
scale better.

\subsection{Conditional Variational Auto-Encoder}
A Conditional Variational Auto-Encoder \cite{kingma2013auto} maximizes
the variational lower bound of the data point $x$ conditioned on
variable $c$. During training, the data point $x$ and its label $c$ are
provided as input to a conditional encoder $q$ that maps the input to
a noise latent space $Z$. Concretely, the conditional encoder learns a
mapping from inputs $(x, c)$ to a distribution $Z$ parameterized by
$\mu$ and diagonal covariance matrix $\Sigma$. During training the
encoder output $z$ is then reparameterized with noise variable
$\epsilon$, $\tilde{z} = z + \epsilon$, via the
\emph{reparameterization trick} \cite{kingma2013auto}, and finally
$(z, c)$ is provided to the decoder $p$ which maps back to image space
$I$. The lower bound $\tilde{\mathcal{L}}_{CVAE}(x, c; \varphi, \phi)$
is thus empirically defined as the sum
\begin{align*}
	-KL(q_\phi(z|x,c)||p_\varphi(z|x)) + \frac{1}{N} \sum_{n=1}^{N}\log p_\varphi(c|x,z).
\end{align*}
\section{Generate and Verify}
The configuration space $C$ is a model of semantically meaningful
features of the operational scenario that should be sufficient to
evaluate the correctness of the system, thus $\dim C$ corresponds to
the number of such features.  For example, in the case
of estimation of the distance $d$ from the camera to a lane marker on
a road from image data produced by a front-mounted vehicle camera,
assuming fixed camera pitch and roll angles, the configuration space
could be the cylinder $S^1 \times [-r,r]$ where $r$ is an estimate of
visual range. Here a point $(\vartheta, d)$ in $C$ corresponds to the
physical scenario in which the camera is at a distance of $d$ meters
from the line oriented with a yaw angle of $\vartheta$ radians with
respect to the line, as depicted in Figure \ref{fig:cspace}.

Given a set of images labeled with the ground truth in our
configuration space, which may be gathered either from the real world
or in simulation, our approach is to learn a deterministic decoder
from configuration space to image space. The decoder may take the form
of a traditional neural network decoder, as in the latter half of an
autoencoder \cite{Hinton2006ReducingTD}, or as a conditional
generative model \cite{Sohn2015LearningSO}. A conditional generative
model allows us to condition our decoder on both our semantically
meaningful configuration space $C$ and a learned noise latent space
$Z$ to capture variation in the image distribution we either do not
know how to, or do not wish to, define semantically. The input space
to be solved over then becomes $C \times Z$. 

\begin{algorithm}\label{algorithm}
	\DontPrintSemicolon % Some LaTeX compilers require you to use 
	% \dontprintsemicolon instead
	\KwData{Target Task Model to be Verified $P_\varphi$, images $X$
		labeled with configuration $c$, reconstruction cost function
		$\mathcal{L}$, partition $(B_{i})$ of $C$, correctness properties
		$\tau_{i}$ with universal quantifiers restricted to the
		corresponding cells $B_{i}$, Neural Network Verification Tool (Solver) $\omega$.}
	Initialize decoder $G_\psi$ with random weights $\psi$\;
	\For{epoch $e\in \{1, \ldots, E\}$} {
		sample random minibatch of training tuples $(x, c)$\;
		$\hat{x} = G_\psi(c)$\;
		Perform a gradient descent step on $\mathcal{L}(x, \hat{x})$.\;
	}
	\For{cell $i \in P$} {
		return $\omega(\tau_i, G_\psi, P_\varphi)$\;			
	}
	\caption{{\sc Generate and Verify}}
	\label{algo:generateandverify}
\end{algorithm}

In what follows, we denote by $P_{\varphi}$ accepting input from image space $I$
parameterized by $\varphi$ and let $G_{\psi}$ denote a neural network
decoder mapping input from configuration space $C$ to image space
$I$ parameterized by $\psi$.  Let $L$ denote a subset of $C$, then
\emph{global correctness modulo the decoder} is the following
statement:
\begin{align}\label{eq:global}
	\forall c \in C. d(P_\varphi(G_\psi(c)), y) < \epsilon
\end{align}
In practice, we carry out the analysis separately over the individual
cells of a \emph{partition} of $C$ as a union of cartesian products of closed
intervals:
\begin{align*}
	C \times Z = \bigcup_i\prod_j[b_{i,j},u_{i,j}]
\end{align*}
Each constituent cartesian product $B_i := \prod_j[b_{i,j},u_{i,j}]$
is referred to as a \emph{cell} of the partition.  We typically
require cells to be almost mutually disjoint in the sense that
$B_i^{\circ} \cap B_k^{\circ} = \emptyset$ for $i\neq k$, where $B_{i}^{\circ}$ denotes the
interior of the set $B_{i}$.

Our algorithm operates in three stages: initialization, training, and solving.  
At the initialization stage, the user specifies a configuration space
$C$, or $C \times Z$, as a list of variables together with their
intended ranges.  

Training the decoder is a standard supervised learning task utilizing
labeled perception data $X$ (e.g., images) such that the labels
correspond to points in $C$. In the case of a conditional decoder,
that is a decoder conditioned on $C$ with learned noise latent $Z$,
the training follows the standard conditional variational auto-encoder
paradigm as previously outlined above (in Section Conditional Variational
Auto-Encoder).

During analysis, a partition, and descriptions of the decoder and
regressor neural network architectures, and their respective weights,
are provided as inputs to our dReal-based neural network analysis tool
summarized in (Section Neural Network Verification above). For each
cell of the partition the analysis outputs one of the following:  
\begin{enumerate}
	\item a mathematical proof of correctness;
	\item a counter-example indicating the presence of a failure case in
	the cell; or
	\item an indication that nothing could be determined regarding the cell.
\end{enumerate}
Here the notion of correctness corresponds to (\ref{eq:global}), but
with the universal quantification restricted to the individual cells
$B_{i}$.  The parameters required are the constant $\varepsilon$ from
(\ref{eq:global}) and the precision $\delta$ used by dReal.

\begin{figure*}[t]
	\centering
	\includegraphics[width=0.9\columnwidth]{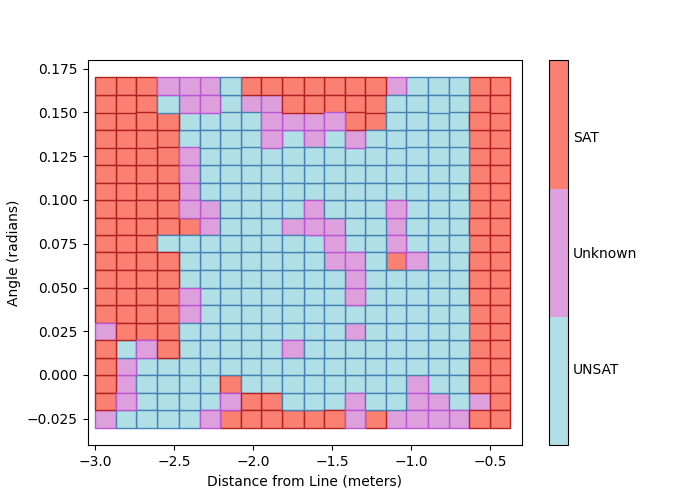} % Reduce the
	% figure size so that it is slightly
	% narrower than the column. Don't use
	% precise values for figure width.This
	% setup will avoid overfull boxes.
	\includegraphics[width=0.9\columnwidth]{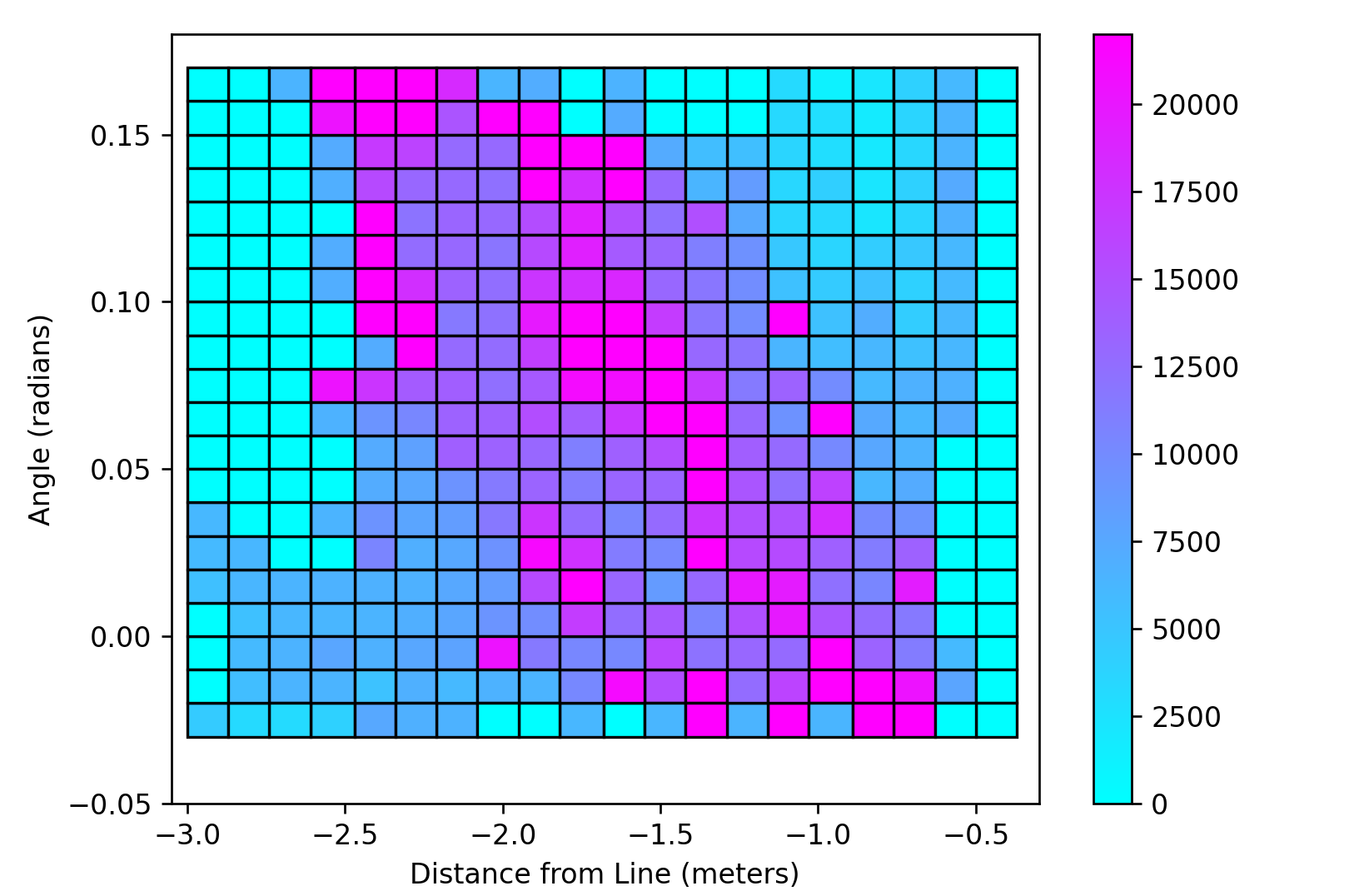} % Reduce the figure size so that it is slightly narrower than the column. Don't use precise values for figure width.This setup will avoid overfull boxes. 
	\caption{Proof map (left) over $C$ space when $\epsilon =
		0.25$. SAT (red) cells contain counterexamples, Unknown
		(purple) cells are inconclusive, and UNSAT (blue)
		cells are provably correct over the infinite set of states
		contained in the interval of state space they
		represent. Heatmap (right) of cell solving times in seconds.}
	\label{fig:proofmap_c}
\end{figure*}

The aggregation of analysis returns over the partition is called a
\emph{proof map}. In the case of a two-dimensional proof map (as is
the case in our first experiment) it can be visualized as a heat map,
as seen in Figure \ref{fig:proofmap_c}. Mathematically, a proof map is
a map $m:I\rightarrow\{-1,0,1\}$ where $I$ is the index set of the
partition. Intuitively, $m(i)=1$ indicates that there exists a
configuration in cell $B_i$ that causes the target model to produce an
incorrect estimate.  Similarly, $m(i)=-1$ indicates that the tool was
able to generate a mathematical proof that, relative to inputs to the
target model generated by the decoder, the target model produces
correct estimates for all configurations in cell $B_i$. Finally,
$m(i)=0$ indicates that the tool was unable to produce a
mathematical proof of correctness for, or to find an example where an
incorrect estimate is produced by a configuration, in cell $B_i$.  In
order to speed up proof map generation some sampling can be
carried out in order to rule out some cells with $m(i)=1$
prior to handing the analysis over to dReal.  In our experiments, we
merely pre-processed by evaluating the centers of cells, but more
nuanced schemes could be deployed.

\section{Experiments}

\begin{figure}[t]
	\centering
	\includegraphics[width=0.9\columnwidth]{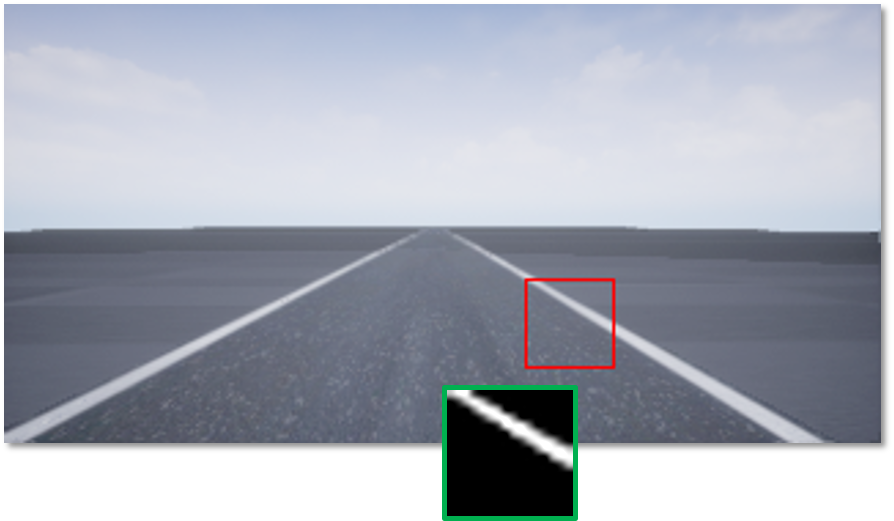}
	\caption{Original camera image from the AirSim simulation
		environment. Crop outlined in red, post processed training
		image inset and outlined in green. View is from the point in
		configuration space ($d=1.5481, \theta=0.0136$).}
	\label{fig:front_center}
\end{figure}

\begin{figure}[t]
	\centering
	\includegraphics[width=0.8\columnwidth]{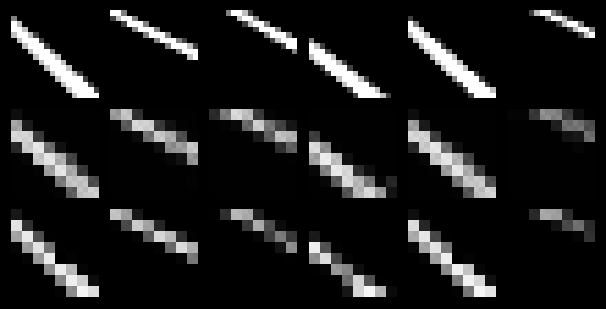}
	\caption{A set of six training images (top), their
		reconstructions (middle), and reconstructions conditioned on $z=0$ (bottom) from six different points in configuration
		space.}
	\label{fig:comparison}
\end{figure}

\begin{figure*}[ht]
	\centering
	\includegraphics[width=0.48\columnwidth]{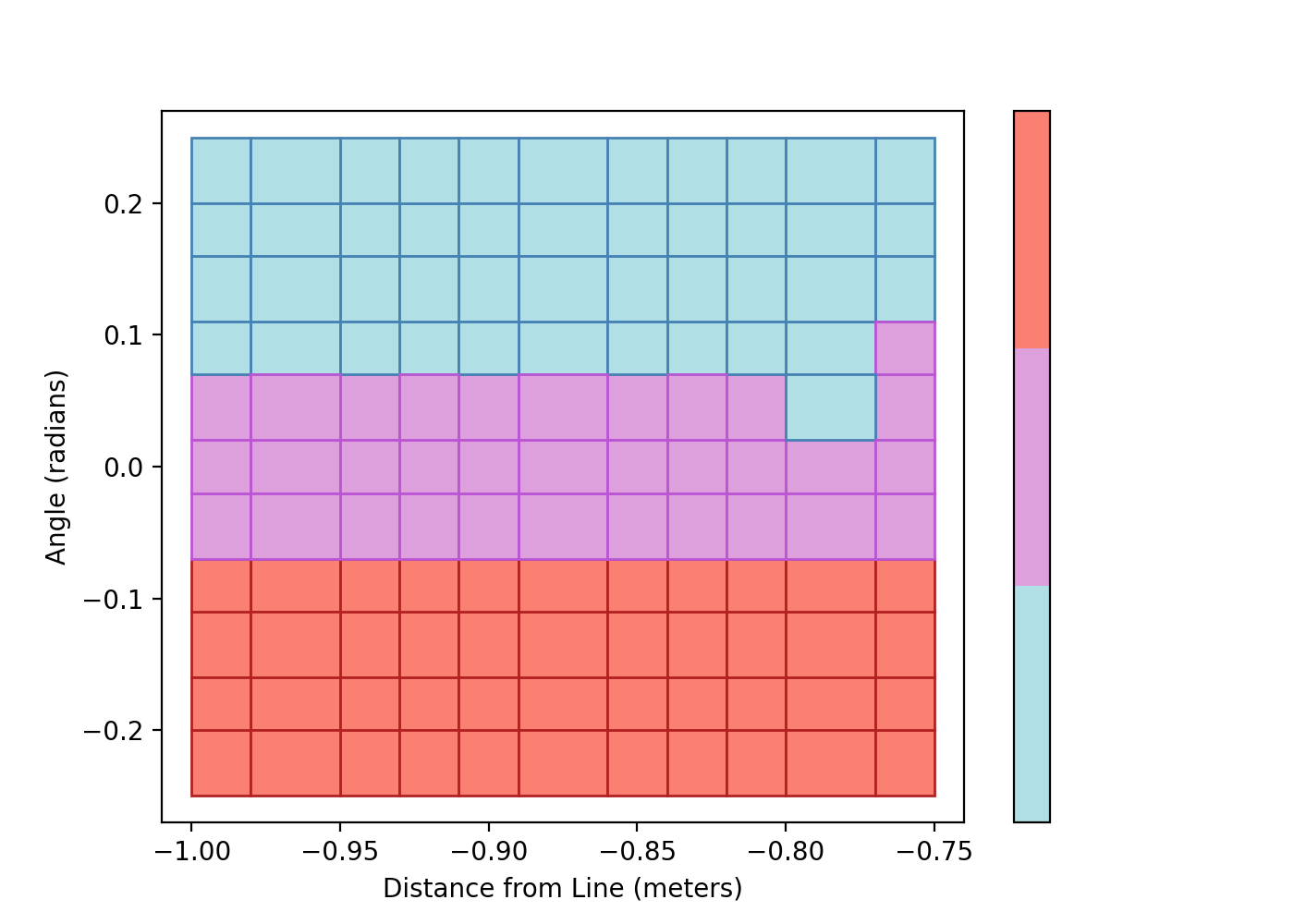}
	\includegraphics[width=0.48\columnwidth]{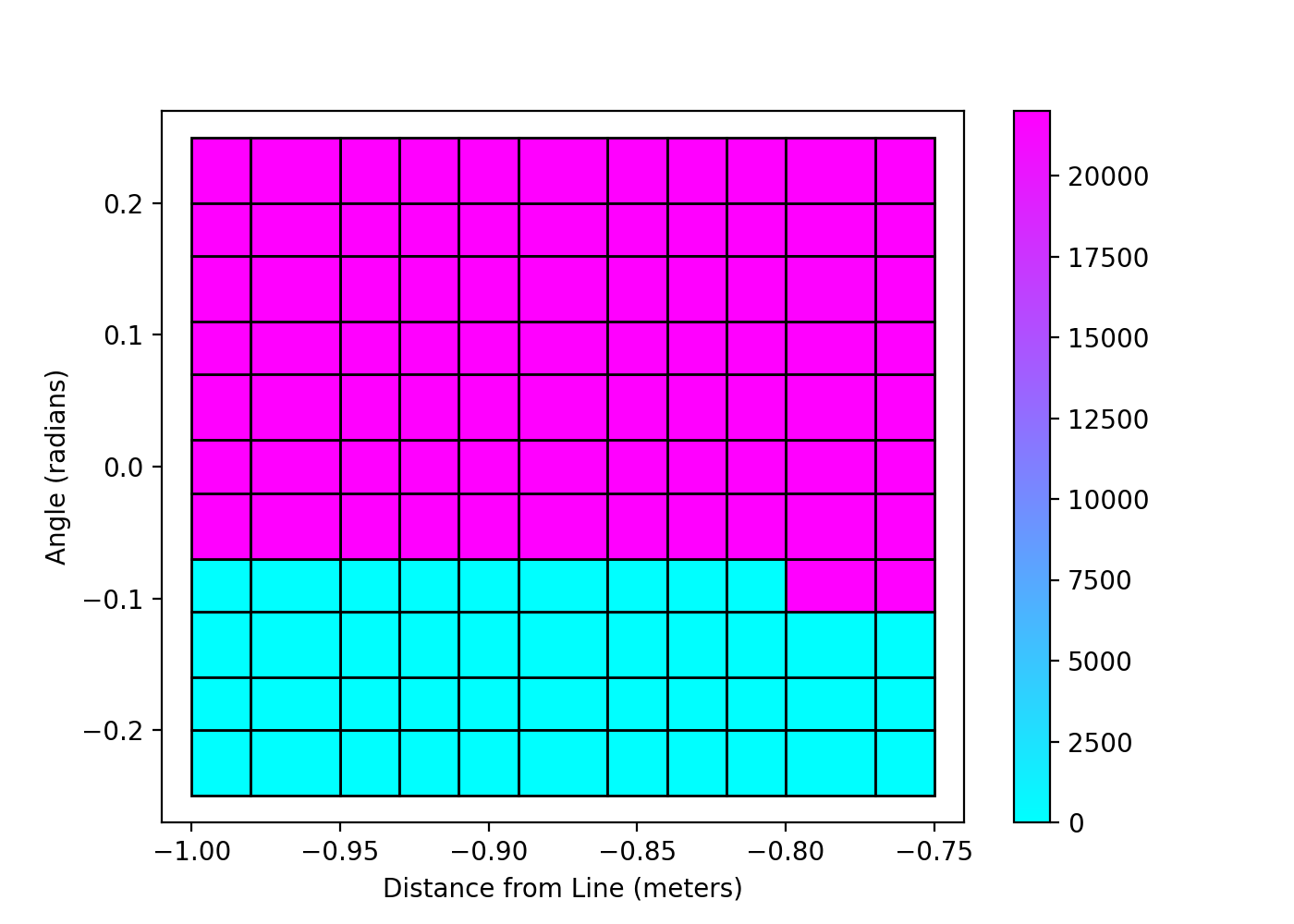}\\
        \includegraphics[width=0.48\columnwidth]{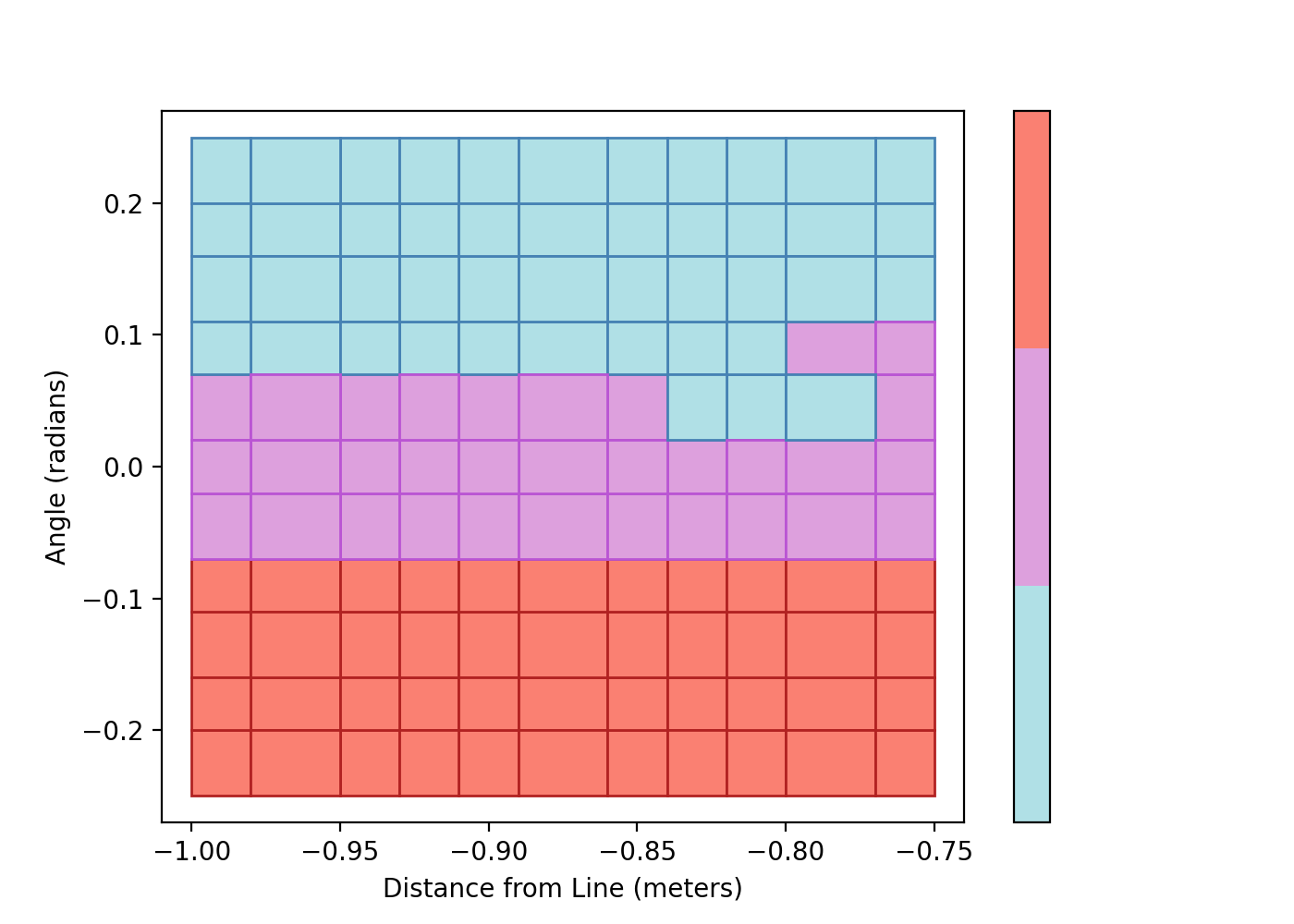}
	\includegraphics[width=0.48\columnwidth]{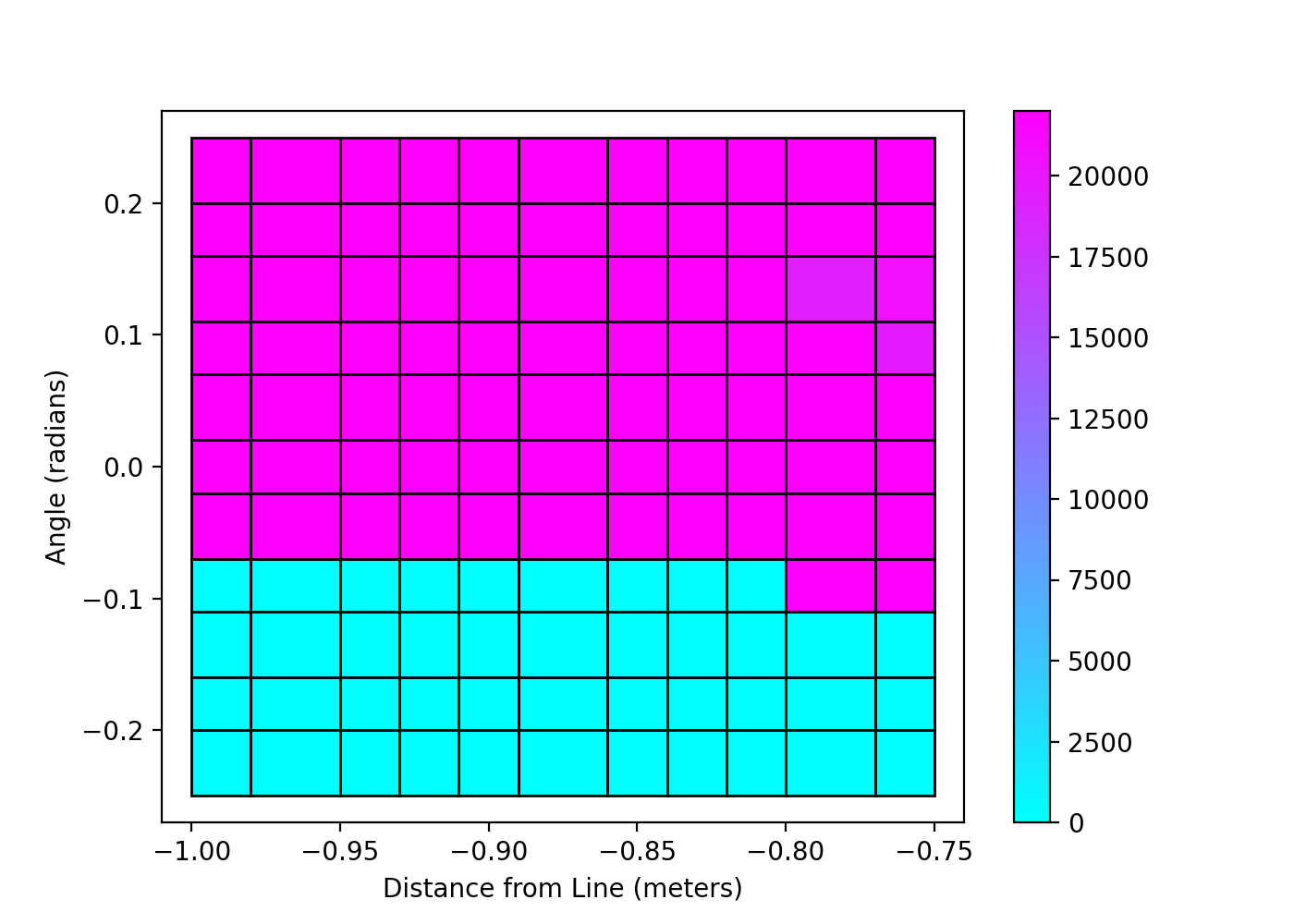}
	\caption{Proof map over $C \times Z$ space when
		$\epsilon = 0.5$ in the latent regions $-0.1\leq z\leq
                0.0$ (top left) and $0.0\leq z\leq 0.1$ (bottom left). SAT (red) cells contain counterexamples,
		Unknown (purple) cells are inconclusive, and UNSAT (blue)
		cells are provably correct over the infinite set of states
		contained in the interval of state space they represent.
		Heatmaps (right) of cell solving times in seconds for the
		same latent regions.}
	\label{fig:proofmap_cz}
\end{figure*}

We will describe our application of the Generate and Verify approach
to a neural network perception system performing regression in two
experiments. In both experiments the task is to estimate the distance
to a lane marker given an image from the front mounted camera of a
vehicle. In the first experiment the configuration space consists only
of two features: the distance in meters to the lane marker $d$, and the
yaw angle of the vehicle $\theta$. A depiction of the configuration
space can be seen in Figure \ref{fig:cspace}. The roll and pitch are
held fixed. In the second experiment an augmentation is randomly
applied to each training image yielding a break in the line of
variable width. This augmentation is not defined in the configuration
space $C$. Instead, it is captured by a traditional Conditional
Variational Auto-Encoder as a learned feature in a latent space $Z$.

The training images are collected in the AirSim
\cite{Shah2017AirSimHV} autonomous driving simulator in a single lane
training environment. After capture, an image is converted to
grayscale, a square crop of the image is taken at a fixed point in the
frame, and the resulting crop is thresholded to yield a black and
white image. Finally the cropped image is resized to $16\times 16$ pixels. The
preprocessing pipeline is depicted in Figure
\ref{fig:front_center}. 10,000 such images are collected in
simulation, along with their $C$ space ground truth labels, to
construct the decoder training set $X$. The structural similarity
index measure (SSIM) between the original images $X$ and their decoder
reconstructions $\hat{X}$ was measured to be $0.823$. Representative
images $x$ and both their decoder and conditional decoder reconstructions $\hat{x}$ from six points in $C$ space are presented in Figure \ref{fig:comparison}. 

Metrics for the evaluation of generative image models are an active area of research and recently the Fr\'echet Inception Distance (FID) \cite{heusel2017gans} and Improved Precision and Recall (P\&R) \cite{kynkaanniemi2019improved} have gained in popularity in the GAN literature. FID measures the Wasserstein-2 distance between two multidimensional Gaussians defined by the mean and covariance of the feature activations of an InceptionV3 classification model for the training images and generated images, respectively. P\&R measures average sample quality ("precision") of generated images and the sample distribution coverage ("recall"). While this work has not incorporated these metrics the generation of realistic images and associated metrics are actively studied and future applications of our method will benefit from advances in this research.

In the first experiment $C$ fully describes the configuration space
and the task of training the decoder is a simple supervised learning
task with input configuration $c$ being mapped to image output
$x$. The network architecture consists of a 2D input and two 64 node
dense hidden layers with ReLU activation. Finally, a 2D transposed
convolution with a single input and output channel, kernel size
$2\times 2$,
and stride 2, with sigmoid activation effectively upsamples the output
of the final dense layer to yield a $16\times 16$ grayscale image. During
training the weights of the decoder are updated such
that they minimize the Binary Cross-Entropy loss between the ground
truth image and the grayscale reconstruction.  

\begin{table}[b]
  \begin{center}
    \begin{tabular}{l|l}
      \textbf{Hyperparameter} & \textbf{Value}\\
      \hline
      batch size & 64\\
      optimizer & Adam\\
      Learning rate & 1e-4\\
      Weight decay & 5e-4\\
      epochs & 500,000\\
    \end{tabular}
    \caption{Decoder and Regressor hyperparameters used during training.}
    \label{tab:hypers}
  \end{center}
\end{table}

\begin{table}[b]
  \begin{center}
    \begin{tabular}{lr}
      \textbf{Network} & \textbf{Loss}\\
      \hline
      Decoder & BCE\\
      Conditional Decoder (CVAE) & BCE + KL Divergence\\
      Regressor & Mean Squared Error\\
    \end{tabular}
    \caption{Decoder and Regressor loss functions used during training.}
    \label{tab:losses}
  \end{center}
\end{table}

The regressor perception network is trained independently on the same
training data as was collected for the decoder. In the first
experiment the regressor architecture consists of a $16\times 16$
pixel single channel image input, followed by an average pooling
downsampling layer with kernel size $2\times 2$, and stride 2. The
pooling layer is followed by a single dense layer of 64 nodes with
hyperbolic tangent activation and a single linear output. The
hyperparameters used during training for both the decoder and
regressor are summarized in Table \ref{tab:hypers}.

In the first experiment analysis, we only considered
$C=[-3,-0.37]\times [-0.03, 0.17]$ and the partition a linearly spaced
$20\times 20$ grid with a total of $400$ cells. In this case, the
precise definition of correctness states that the estimates are not too far from the
ground-truth distance in the sense that:
\begin{align*}
	\forall (d,\vartheta) \in B_i. \|P_\varphi(G_\psi(d,\vartheta)) - d\| < \epsilon
\end{align*}
where the precision constant $\epsilon$ is 0.25 and $\|\cdot\|$
denotes the $L_{2}$ norm.

In the second experiment, we consider $C=[-1., -0.75]\times[-0.25,
0.2]$ and $Z=[-0.1,0.1]$ with a linear partition of size two along the
$Z$ axis and a linearly spaced $11\times 11$ grid along $C$.  We also
evaluated separately the same grid over $C$ with the latent $Z$ region
$[0.2,0.3]$.  In this experiment we used $\epsilon=0.5$.

\subsection{Decoder Results}
\begin{table}[t]
  \centering
  \begin{tabular}{lcc}
    \textbf{Cell Result} & \textbf{Count} & \textbf{Percent} \\
    \hline
    UNSAT & 218 & 55\% \\
    SAT & 56 & 14\% \\
    Unknown & 126 & 32\% \\
  \end{tabular}
  \caption{Aggregated $C$ space cell results.}
  \label{tab:c_results}
\end{table}

\begin{table}[t]
  \centering
  \begin{tabular}{lcc}
    \textbf{Cell Result} & \textbf{Count} & \textbf{Percent} \\
    \hline
    UNSAT & 89 & 37\% \\
    SAT & 88 & 36\% \\
    Unknown & 65 & 27\% \\
  \end{tabular}
  \caption{Aggregated $C \times Z$ space cell results.}
  \label{tab:cz_results}
\end{table}
Figure \ref{fig:proofmap_c} shows the proof map corresponding to our
first experiment resulting from solving
over the portion of $C$ space with distance $d \in [-3, 0]$ (negative
values correspond to being left of the lane marker) and yaw angle
$\theta \in [-.1, .2]$. The property proved is that the the regressor
will produce estimates for each point $c$ within .25m of the ground
truth. Counter-examples are found when the vehicle is near the lane
marker at negative yaw values, and when distant from the lane marker
(near 3 meters) at positive yaw angles. The regressor is found to be
provably correct over 55\% of the cells of the partition, with each
cell $b_i$ representing the infinite set of images the decoder could
generate from the infinite set of points $c$ within each
cell. Crucially, the correct cells are not an approximation of the
regressor's accuracy with respect to the decoder in those regions of
$C$ space, as one would acquire with traditional sample and test
techniques, rather they are regions in which the solver returns a
mathematical proof that the regressor will return estimates within
.25m of ground truth $\forall c \in B_i$. Summary statistics are
provided in \ref{tab:c_results}.

The total time taken to solve over the proof map partition in
\ref{fig:proofmap_c} was 208.43 hours, approximately 8.5 days. Each
cell is solved by a separate process and a maximum of fives cells are
solved simultaneously. A heatmap of the time taken for each cell to
return a result is visualized in Figure \ref{fig:proofmap_c}. It is
apparent that counter-examples are returned more quickly than proofs
of correctness. Further timing statistics and details regarding the
machines on which our experiments were carried out are included in the
appendix.

\subsection{Conditional Decoder Results}

In the second experiment with equal probability a diagonal mask is
applied to each training image yielding a break in the line of
variable width, thus capturing the notion of one way in which our lane
marker may be degraded in the real world. The configuration space $C$
is identical to the first experiment, but we now concatenate it with
the learned latent feature space $Z$. $Z$, and the decoder, are
learned simultaneously in a CVAE architecture. The CVAE network
architecture consists of encoder network $q$ with input layer
dimension ($16\times 16)+\dim C$, followed by densely connected layers with
ReLU activation of 8 nodes and 2 nodes respectively, and finally
single linear outputs for each of the $\mu$ and $\sigma$ parameters of
the latent space $Z$. The decoder architecture remained unaltered from
the first experiment. 

It should be noted that in order to maintain independence between the
information in $C$ and $Z$, that is, in order for the decoder output
to remain fixed at a point $c$ while $z$ is varied, it was found that
the capacity of the encoder needed to be limited. Now, by simply
varying the values of our input $z$ we can generate images of the
entire range of lane marker degradation our learned model has
captured, conditioned on our fixed point in $C$ space.  

The architecture of the regressor from the first experiment was found
to be insufficient to handle the noise present in the images of the
second experiment so a revised architecture was developed. The
regressor in the second experiment consists of $16\times 16$ pixel single
channel image input layer, followed by an average pooling layer with
kernel size $2\times 2$ and stride 2. The downsampling pooling layer is
followed by a 2D convolutional layer with single channel input and
output, kernel size $3\times 3$, stride of 1, no padding, and hyperbolic
tangent activation. A dense layer of 32 nodes with hyperbolic tangent
activation follows the convolutional layer before a final single
output node with sigmoid activation provides scaled estimates of the
distance $d$.

Proof maps, for latent regions $[-0.1,0]$ and $[0,0.1]$, and timing
heatmaps can be found in Figure \ref{fig:proofmap_cz}.  The result
statistics are in Table \ref{tab:cz_results}. 

\section{Conclusion}
We have described a novel approach to neural network evaluation that
allows a more fine-grained analysis than state-of-the-art testing and
simulation approaches: it allows us to, via mathematical analysis,
survey an infinite set of possible scenarios/images, whereas
testing/simulation necessarily only examines a finite set of such
scenarios/images. Our approach complements that of traditional sample
and test methods by allowing stronger assurance guarantees over
infinite (non-measure zero) sets of images and active counter-example
search, counter-examples that might otherwise only be found in
production as \emph{edge cases}. These counter-examples are
particularly valuable in debugging and developing insight into our
models and data as they tell us under exactly what inputs our model
will fail. These counterexamples can lead us to the discovery of
deficiencies in our training data and patterns of success and failure that
may be attributable to strengths and defects in our model design. Our
generative decoder can be further utilized to generate training data
around failure cases, allowing us to both find and resolve so called edge case
failures before ever encountering them through traditional testing or
in production. Our training data was captured at a fixed camera rotation step size and our simple decoder is able to smoothly interpolate between view angles to generate correct intermediate views absent from the training data, and to correctly capture the relationship between camera distance and angle to the lane marker. These novel views are not captured in the dataset, and thus could not be tested against with traditional testing schemes. Further, we can capture additional data variation, and generate images with that variation that were not present in the data: our CVAE is able to generate images with a "line break" augmentation in a smooth range of no break, and narrow to wide break, from every viewing angle, despite seeing only a finite set of training data that did not capture this full variation.
As we refine our models in production, our approach
provides a way to verify that we have not introduced new failure cases
into previously correct task performance
\cite{mccloskey1989catastrophic}. Additionally, there are several
important differences between our approach and state-of-the-art
approaches in the verification of neural networks. In the literature
on verification of neural networks the focus is always on local
properties, in the sense described previously, and in particular on
adversarial robustness properties (i.e., that the system will not
behave differently when inputs are subject to small
perturbations/disturbances). We address global notions of correctness
similar to those from \cite{Fijalkow2019VerificationON}, but where our
notion is modulo a generative model with input space that has a
semantically meaningful factor.  This approach can help to answer
questions of the form: are there physical configurations of objects
that will give rise to incorrect estimates by the system. This link
between the actual physical configuration and perception network
correctness is entirely broken in the existing work on verification of neural
networks. Our contribution is thus a novel perspective on the
verification of correct behavior in neural network perception systems.

The principal drawback of our approach is the fact that current neural
network verification tools scale poorly to large networks or apply
only to those networks having prescribed architecture features.  This
limits the size of both the models we can analyze and the generative
models that we can employ in the analysis.  Nonetheless, we believe
that future advances will make this approach increasingly tractable
in general.  In the mean time, our approach can furthermore be used
to augment testing and adversarial machine learning based approaches
to neural network robustness analysis. 

\section{ Acknowledgments}
We would like to thank Doug Stuart, Ramesh S., Huafeng Yu, Sicun Gao,
and Aleksey Nogin for useful conversations on topics related to this
paper.  We are also grateful to Cem Saraydar, Tom Bui, Chad McFarland,
Bala Chidambaram, Roy Matic, Mike Daily and Son Dao for their support
of and guidance regarding this research.

\appendix
\section{Machine Information and Hyperparameter Tables}
The machine learning experiments were carried out on a machine with
Intel Xeon E5-2680 v4 @ 2.40GHz processor and two NVIDIA Tesla P100-SXM2-16GB GPUs.  The neural
network verification was done on a machine with Intel Xeon
E5-2690 v3 at 2.60GHz processor, 24 CPU cores, and 1TB of memory.

\begin{table}[h]
  \begin{center}
    \begin{tabular}{l|l}
      \textbf{Hyperparameter} & \textbf{Value}\\
      \hline
      batch size & 64\\
      optimizer & Adam\\
      Learning rate & 1e-4\\
      Weight decay & 5e-4\\
      epochs & 500,000\\
    \end{tabular}
    \caption{Decoder and Regressor hyperparameters used during training.}
    \label{tab:hypers}
  \end{center}
\end{table}

\begin{table}[h]
  \begin{center}
    \begin{tabular}{l|l}
      \textbf{Network} & \textbf{Loss}\\
      \hline
      Decoder & Binary Cross Entropy (BCE)\\
      Conditional Decoder (CVAE) & BCE + KL Divergence\\
      Regressor & Mean Squared Error\\
    \end{tabular}
    \caption{Decoder and Regressor loss functions used during training.}
    \label{tab:losses}
  \end{center}
\end{table}

\begin{figure}[t]
  \centering
  \includegraphics[width=0.8\columnwidth]{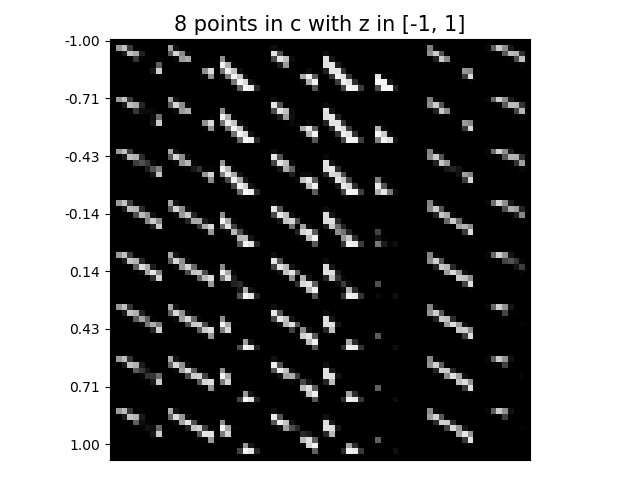}
  \caption{Points in $C$ while varying $z$. The perspective defined by $c$ remains fixed while $z$ varies the line break perturbation.}
  \label{fig:z_comparison}
\end{figure}

\bibliographystyle{plain}
\bibliography{GenerateVerify}
\end{document}